\documentclass[a4paper,11pt]{article}
\usepackage[margin=1in]{geometry}
\ifdefined\XeTeXgenerateactualtext
\usepackage{fontspec}
\newfontfamily\chinesefont{FandolSong-Regular.otf}[
  BoldFont=FandolSong-Regular.otf,
  ItalicFont=FandolSong-Regular.otf,
  BoldItalicFont=FandolSong-Regular.otf]
\newcommand{\zh}[1]{{\chinesefont\XeTeXlinebreaklocale "zh"\relax
  \XeTeXlinebreakskip=0pt plus .12pt\relax #1}}
\else
\newcommand{\zh}[1]{#1}
\fi
\usepackage{amsmath,graphicx,array,booktabs,longtable,calc,titlesec}
\usepackage[unicode,hidelinks]{hyperref}
\titleformat{\section}{\fontsize{14}{17}\selectfont\bfseries}{}{0pt}{}
\titleformat{\subsection}{\normalsize\bfseries}{}{0pt}{}
\titlespacing*{\section}{0pt}{15pt plus 2pt}{5pt}
\titlespacing*{\subsection}{0pt}{12pt plus 2pt}{3pt}

\newcommand{\papergraphic}[2]{%
  \IfFileExists{figures/#1.pdf}{%
    \includegraphics[#2]{figures/#1.pdf}%
  }{%
    \includegraphics[#2]{figures/#1.png}%
  }%
}
\ifdefined\XeTeXgenerateactualtext
\newcounter{paperfigure}
\newcommand{\paperfigure}[3]{%
  \par\addvspace{8pt}\noindent
  \begin{minipage}{\linewidth}\centering
    \papergraphic{#1}{width=#2,height=.58\textheight,keepaspectratio}%
    \par\vspace{4pt}\small\refstepcounter{paperfigure}%
    Fig~\thepaperfigure.\ #3\par
  \end{minipage}\par\addvspace{8pt}%
}
\else
\newcommand{\paperfigure}[3]{%
  \begin{figure}
    \centering
    \papergraphic{#1}{width=1.5\linewidth}%
    \caption{#3}%
  \end{figure}%
}
\fi

\newcommand{\papertitle}{An Interpretable Stylometric Framework for Early Chinese Buddhist Translator Styles: A Case Study on the Attribution of Disputed Works of An Shigao}
\title{\papertitle}
\author{Yating Pan\textsuperscript{1,3} \and
  Xinyue Zhang\textsuperscript{2} \and Jun Wang\textsuperscript{1,3}}
\date{}
\hypersetup{
  pdftitle={\papertitle},
  pdfauthor={Yating Pan, Xinyue Zhang, Jun Wang}
}
\newcommand{\paperdetails}{%
  {\small
  \textsuperscript{1}Department of Information Management, Peking University\\
  \textsuperscript{2}Department of Philosophy and Religious Studies, Peking University\\
  \textsuperscript{3}Research Center for Digital Humanities, Peking University\par}
  \vspace{7pt}
  {\small
  Presented as a long paper at Digital Humanities 2026 (DH2026).\\
  The conference abstract appears in the revised
  \href{https://doi.org/10.5281/zenodo.22656106}{DH2026 Book of Abstracts},
  pp.~545--549.\par}%
}

\ifdefined\XeTeXgenerateactualtext
\renewcommand{\maketitle}{%
  \begin{center}
    {\fontsize{14}{19}\selectfont\bfseries\papertitle\par}
    \vspace{5pt}
    Yating Pan\textsuperscript{1,3}\enspace
    Xinyue Zhang\textsuperscript{2}\enspace
    Jun Wang\textsuperscript{1,3}\par
    \vspace{3pt}
    \paperdetails
  \end{center}
  \vspace{8pt}%
}
\fi

\begin{document}
\maketitle
\ifdefined\XeTeXgenerateactualtext
\else
\begin{center}
  \paperdetails
\end{center}
\fi

\section{Introduction}

The history of Chinese Buddhist translation (2nd--10th c.) marks a pivotal stage in cultural exchange among India, Central Asia, and China (Zürcher, 1991; Nattier, 2008), yet how translators' stylistic choices evolved remains underexplored. Traditional philological research, from Zürcher (1991) to Hung, Hsieh, and Bingenheimer (2017), identified translator-specific traits but lacked systematic quantification. Recent computational work has advanced this field: Hung et al. (2017) applied PCA to trace diachronic stylistic change; Bingenheimer (2020) used text mining for inter-Āgama relations; and Lu et al. (2025) employed BERT for attribution with 95.2\% accuracy.

Although deep models like BERT achieve strong results, they remain black boxes for philological argumentation. In contrast, traditional approaches such as stylometry perform well in classical languages (Gianitsos et al., 2019) and remain widely used in digital humanities for their interpretability (Holmes, 1998). We therefore adopt a domain-expert-guided, interpretable stylometric framework that extracts linguistically grounded features from CBETA scriptures (CBETA, 2025) and statistically validates them. Applied to disputed works of An Shigao, the framework produces results consistent with both philological scholarship (Nattier, 2008; Zacchetti, 2010) and BERT-based models, demonstrating that interpretable stylometry can rival deep learning while remaining transparent and reproducible, without any supervised training.

\section{Data Collection and Preprocessing}

We compile a CBETA-based corpus of authenticated translations by seven early translators from Eastern Han to Western Jin: Yan Fotiao, An Shigao, Kang Senghui, Kang Mengxiang, Lokakṣema, Zhi Qian, and Dharmarakṣa. The set comprises 31 texts, providing stylistic baselines for early translation schools (Appendix A, Table A1). A secondary evaluation set includes eight disputed works historically associated with An Shigao in classical catalogs but questioned or reassessed by modern philology (Appendix A, Table A2).

Preprocessing removes titles and annotations, retains CJK characters, segments sentences using CBETA punctuation, and preserves intra-sentence punctuation to maintain sentence-pattern signals. All analyses are performed at the document level with subsequent aggregation by translator and period.

\section{Methods}

Figure 1 outlines our expert-in-the-loop pipeline. A Buddhist philology expert defines and validates linguistic feature families, while an NLP engineer extracts these features from authenticated and disputed translations to build author and document style profiles. The classification module then compares them to produce attribution results.

\paperfigure{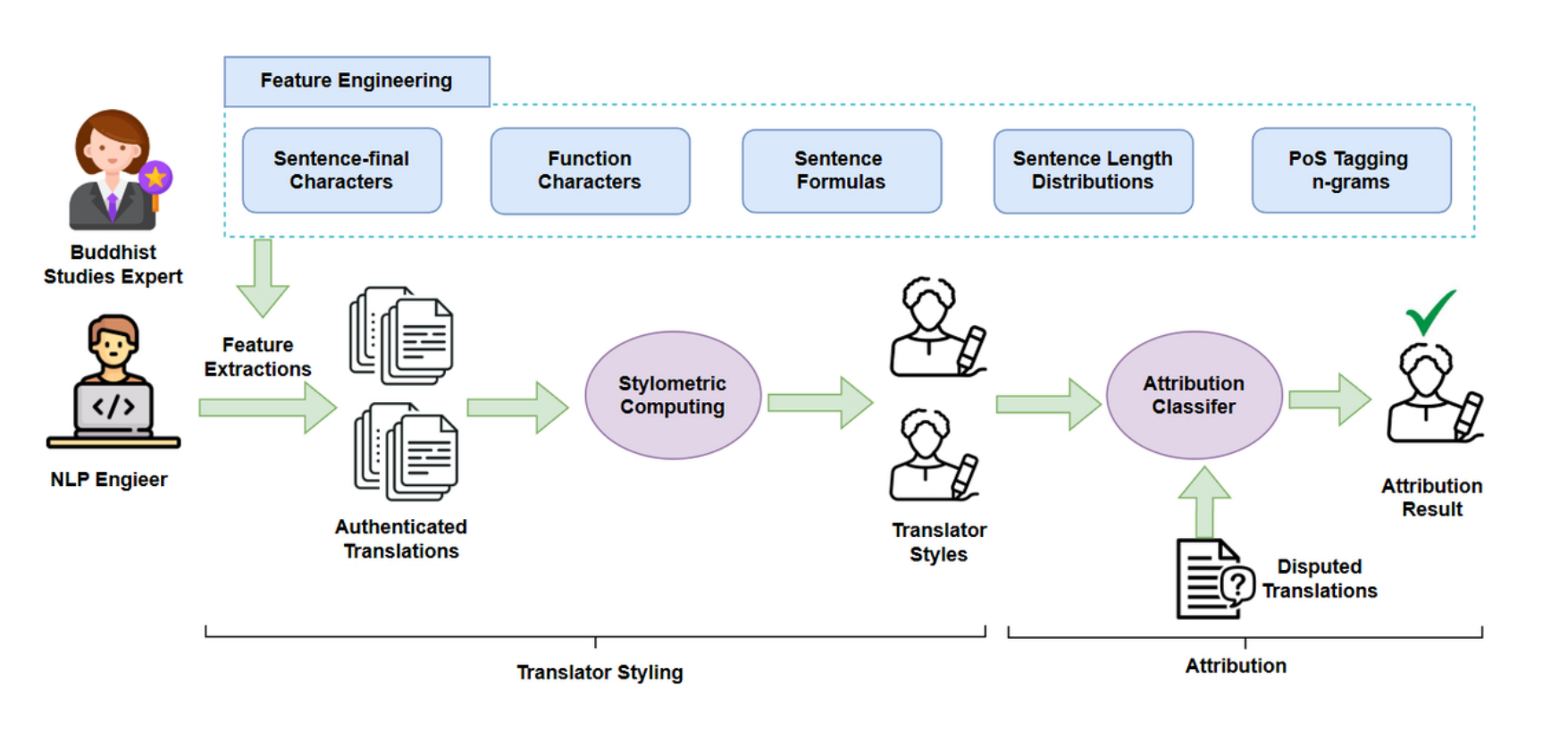}{.91\linewidth}{Overview of the expert-in-the-loop stylometry pipeline.}

\subsection{Feature Engineering with Expert Validation}

We constructed five interpretable stylometric feature families. Two are expert-defined (function words; sentence-formula templates). Three are data-driven (sentence-final characters; sentence-length; POS n-grams) and expert-verified. Feature details are shown in Appendix B, Tables B1-B4.

Let a document have \(S\) sentences (\(s_{1}\),...,\(s_{S}\)). Define total CJK characters \(T = \sum_{i = 1}^{S}\left| \text{CJK}\left( s_{i} \right) \right|\). Define POS tag sequence \(P = \left( p_{1},\ldots,p_{L} \right)\).

\textbf{Function words:} Philology experts defined twenty function words covering pronouns, prepositions, particles, adverbs and conjunctions. For each function word \(func_{w}\), we computed its frequency per 100 Chinese characters:

\[\mathrm{func}_{w,\mathrm{per100}} = 100 \times \frac{\operatorname{count}\left( w\text{ in }\bigoplus_{i = 1}^{S}{\text{CJK}\left( s_{i} \right)} \right)}{T}\]

\textbf{Sentence formulas:} Philology experts defined six rhetorical sentence formulas (enumerative, definitional, categorical, interrogative-enumerative, direct-quotation, and conditional-temporal) to encode the translators\textquotesingle{} personal pedagogical and exegetical expressions. For each formula \(f\), let \(1_{f}\left( s_{i} \right) = 1\) if sentence \(s_{i}\) matches \(f\) once. A sentence may match multiple templates independently, so we computed each pattern\textquotesingle s proportion out of the total sentence count:

\[\text{pattern}_{f}^{prop} = \frac{1}{S}\sum_{i = 1}^{S}{I_{f}\left( s_{i} \right)}\]

\textbf{Sentence-final characters:} We mined top-10 sentence-final characters that most frequently appear across more than half translators to reflect their habitual closure styles. For each character \(c\), we computed its frequency among total sentence counts:

\[\mathrm{final\_prop}_{c} = \frac{\text{\#}\left\{ i \in \text{\{}1,\ldots,S\text{\}}\mid\text{lastCJK}\left( s_{i} \right) = c \right\}}{S}\]

\textbf{Sentence-length distribution:} We computed the mean, median and standard deviation of sentence length for each document to quantify a translator\textquotesingle s structural rhythm and rhetorical density. Short, uniform sentences signal segmented oral prose, whereas long and connective elaborations indicate a shift toward literati styles.

\textbf{POS n-grams:} Part-of-speech n-grams capture stylistic consistency in local syntactic dependencies. We used the Jiayan Toolkit to build a global vocabulary of top-k POS n-grams (n=1..4) from the corpus. For each \(n\), we computed the relative frequency of every vocabulary n-gram pattern \(g\) (e.g., \(g = n_{v}\), \(g = v_{v_{u}}\), etc) within a document\textquotesingle s POS tag sequence \(P = \left( p_{1},\ldots,p_{L} \right)\):

\[
\begin{aligned}
\mathrm{pos}n\mathrm{\_}g\mathrm{\_prop} &= \frac{\operatorname{count}(g\text{ in }P)}{N_n},\\
\mathrm{pos}n\mathrm{\_N} &= N_n,\\
\mathrm{pos}n\mathrm{\_covered} &= \#\{g\in\mathcal{V}_n \mid \operatorname{count}(g)>0\}.
\end{aligned}
\]

\subsection{Evidence-Enhanced Translator Styling}

We grouped all features by translator and applied the Kruskal--Wallis (H) test with Holm--Bonferroni correction \(\left( p_{holm} < 0.01 \right)\). The ten most discriminative features (Appendix B, Table B4) were retained, and standardized scores \(\left( z_{f} = \left( x_{f} - \mu_{f} \right)\text{/}\sigma_{f} \right)\) were computed to build normalized stylistic fingerprints. Translator profiles were then mapped to 0--100 indices via corpus-wise z-scaling and sigmoid transformation, with pooled standard deviations stabilizing small-sample variation. Results across translations and translators are in Appendix C Table C1 and C2.

\subsection{Attribution Scoring for Disputed Translations}

Each disputed text was transformed into the same sigmoid-normalized space as the author profiles by z-scoring each feature (using global \ensuremath{\mu} and \ensuremath{\sigma} from non-disputed texts) and applying a sigmoid. Similarity was then computed as the inverse of a standardized chi-square distance across features, yielding \(\mathrm{Sim}\in[0,1]\). We interpret Sim comparatively, with \(\mathrm{Sim}\geq 0.60\) indicating high confidence.

\section{Results and Discussion}

\subsection{Translator Stylometric Differentiation}

Figure 2 visualizes the ten core stylometric features across seven translators stratified by historical period, revealing the interconnection between individual translator style and Buddhist translation norms. The earliest Eastern Han translators, An Shigao and Yan Fotiao, exemplify literal, plain, and colloquial translation, marked by extremely low function word density and minimal 4-grams. Notably, An Shigao\textquotesingle s many long sentences with over 40 characters contain no function words (T0031: \zh{以应念法不念，不应念法者便念，便爱流生，已爱流生便增多不致，未生爱流亦痴流便生，已生爱流亦痴流便增多不致}). Instead, his Verb-Adverb 2-gram (70.7\%) and Adverb-Verb-Adverb 3-gram scores (71.9\%) dominate, exemplified by the exhortatory "\zh{为}+\zh{不}+verb" pattern (T0014: \zh{一为不恭敬佛、二为不恭敬法、三为不恭敬同学者、四为不恭敬戒}), reflecting the Eastern Han\textquotesingle s oral transmission tradition.

In contrast, Lokakṣema's function words increase (\zh{之}: 44.7\%, \zh{其}: 75.1\%), signaling a shift toward formal expression. By the Three Kingdoms period, translators like Zhi Qian and Kang Senghui show systematic growth in connective density (\zh{之}: 66.4\% and 87.6\%), replacing literal translation with natural cohesion through complex nominal modification (Zhi Qian T0054: \zh{有淫之意、有怒之意、有痴之意}; Kang Senghui T0152: \zh{今又犯之，种无量之罪}).

Dharmarakṣa represents the apex, displaying a mature written style. His elevated function word integration (\zh{之}: 72.6\%, \zh{及}: 81.8\%) and higher-order structures (pos4\_n\_v\_r\_n: 71.5\%) exemplify Noun-Verb-Pronoun-Noun subordination (T0154: \zh{菩萨入之心，释子告其人}), demonstrating Buddhist sutra integration into China\textquotesingle s literary tradition. This evidence traces the sinicization trajectory from the Eastern Han\textquotesingle s plain vernacular to the Western Jin\textquotesingle s refined written expression (Hung et al., 2017).

\paperfigure{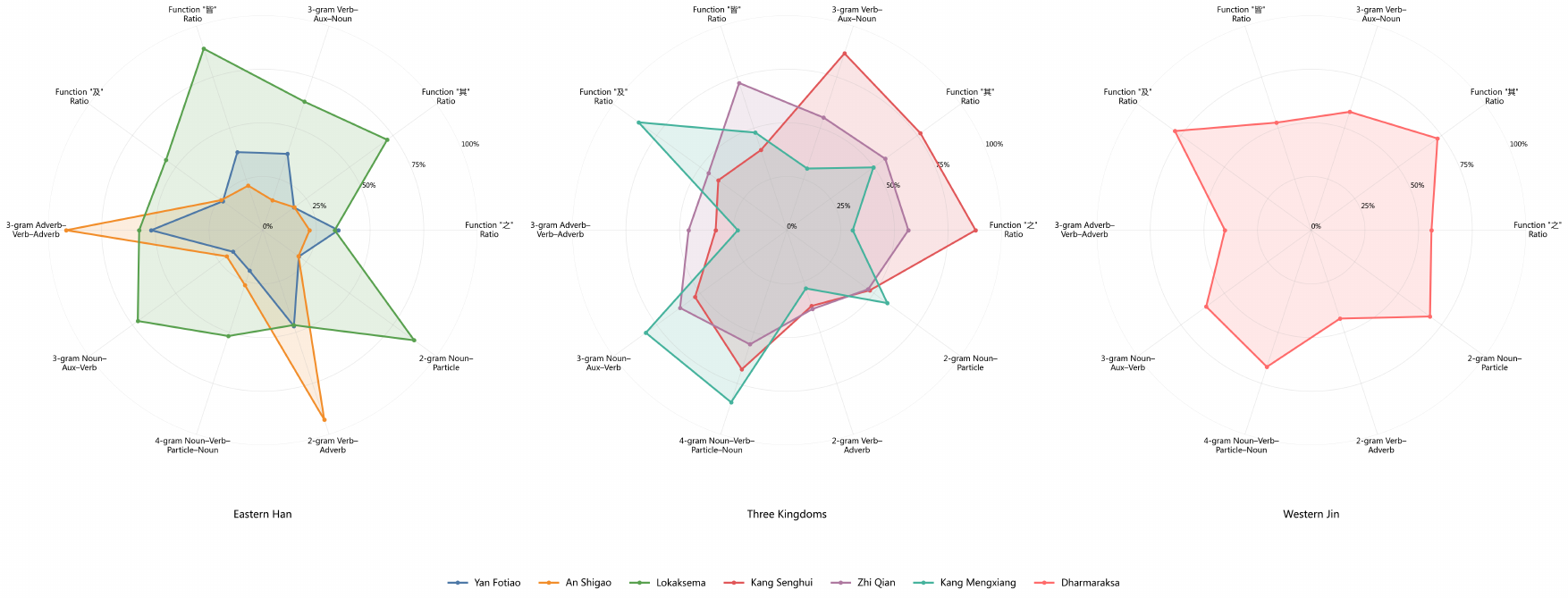}{.94\linewidth}{Stylometric radar charts of early Chinese Buddhist translators based on the top-10 features.}

\subsection{Case Study: An Shigao\textquotesingle s Disputed Translation Attribution}

\paperfigure{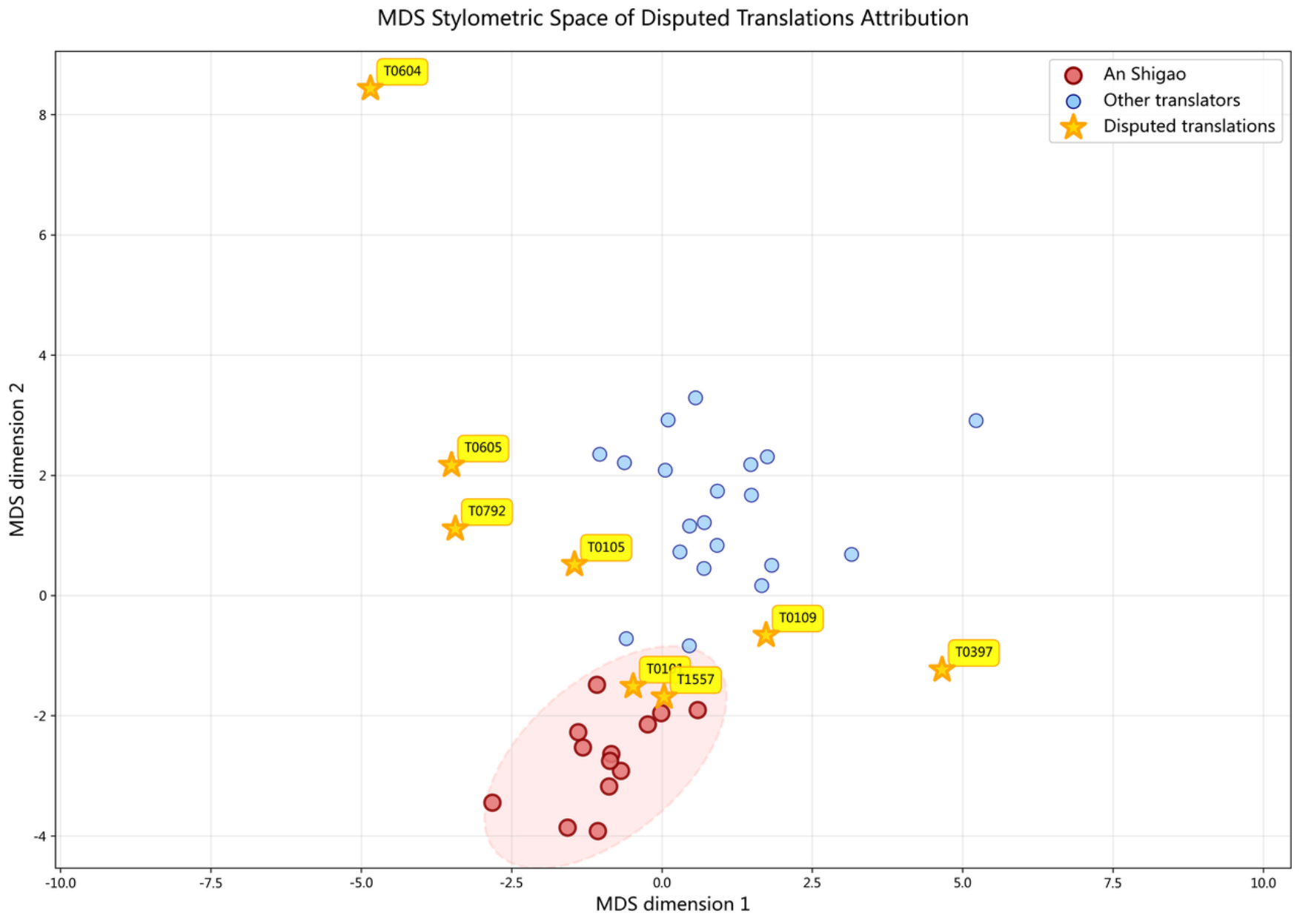}{.85\linewidth}{Stylistic space of early Chinese Buddhist translations in multidimensional spaces.}

\paperfigure{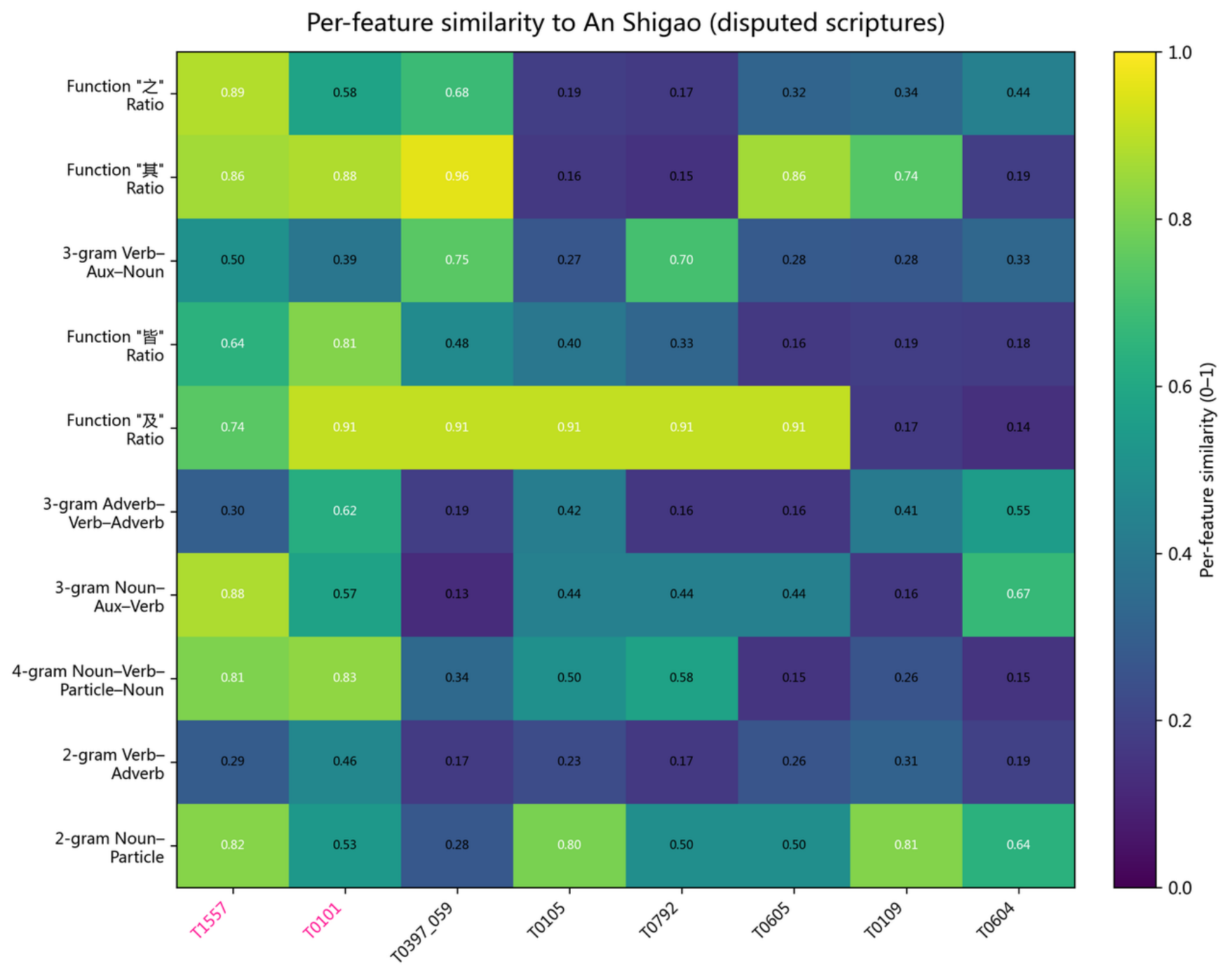}{.79\linewidth}{Similarity distribution between disputed translations and An Shigao\textquotesingle s stylometric styles}

Our attribution results reveal two clear stylistic groups among the disputed translations. T1557 (Abhidharma Prakaraṇa Sūtra) and T0101 (Saṃyuktāgama Sūtra) cluster closely with the verified An Shigao corpus, whereas five other texts diverge sharply. As shown in Figure 3, the red ellipse marks the centroid and 95 \% confidence interval of authentic An Shigao works; blue dots denote other translators from the Eastern Han to Western Jin; yellow stars represent disputed texts. T1557 lies at the center of the An Shigao cluster (Sim = 0.783), displaying his hallmark concise syntax, balanced sentence-formula use, and restrained function-word frequency, consistent with Nattier (2008) and Zacchetti (2010). T0101 (Sim = 0.687) also falls within this range, sharing the rhythmic compactness and pedagogical tone described by Lin (2010), Harrison (2002), and Hayashiya (1941). In contrast, T0109, T0105, T0605, T0792, and T0604 lie far outside the An Shigao region (\(\mathrm{Sim}<0.3\)), featuring poetic passages and lexical registers characteristic of other or even later translators (Nattier 2008; Zürcher 1991; Hayashiya 1941).

Figure 4 visualizes feature-level similarity across the ten core stylometric indicators. Both T1557 and T0101 show high correspondence with An Shigao's function-word ratios and 4-gram density, providing linguistic validation of their authenticity. These findings also align with Lu et al. (2025)'s BERT-based attribution model, yet our zero-training, expert-interpretable framework offers transparent, feature-based explanations suitable for philological inquiry.

\section{Conclusion}

We presented a domain-expert-guided, fully interpretable stylometric framework for early Chinese Buddhist translation studies. Using expert-guided linguistic features and stylometric statistics, the model reconstructs stylistic evolution across the Eastern Han--Western Jin and attributes disputed texts without task-specific training. In the case of An Shigao\textquotesingle s disputed translation attribution, our results match the advanced deep learning method results. Unlike neural classifiers that require labeled training and offer limited rationale, our pipeline attains comparable results while returning human-interpretable evidence.

We note some practical limitations affecting the interpretation. Shorter texts (e.g., T0604, T0605) have fewer features and lower feature stability; currently, the focus is mainly on early Chinese translations of Buddhist scriptures. We will expand to longer corpora and later translators (e.g., Kumarajiva and Xuanzang from the Tang Dynasty), and publish a catalog associated with the evidence to facilitate reproduction. Combining our interpretation with neural models is a promising direction for hybrid, interpretable digital humanities.

\section{Bibliography}
CBETA. (2025). \emph{Chinese Buddhist Electronic Text Association (CBETA) digital corpus} {[}Dataset{]}. \href{https://cbeta.org}{https://cbeta.org}

Devlin, J., Chang, M.-W., Lee, K., \& Toutanova, K. (2019). BERT: Pre-training of deep bidirectional transformers for language understanding. \emph{arXiv:1810.04805.} \href{https://arxiv.org/abs/1810.04805}{https://arxiv.org/abs/1810.04805}

Gianitsos, E., Bolt, T., Chaudhuri, P., \& Dexter, J. P. (2019). Stylometric Classification of Ancient Greek Literary Texts by Genre. In \emph{Proceedings of LaTeCH-CLfL 2019} (pp. 52--60). Association for Computational Linguistics. \url{https://doi.org/10.18653/v1/W19-2507}

Holmes, D. I. (1998). The Evolution of Stylometry in Humanities Scholarship. \emph{Digital Scholarship in the Humanities} (formerly \emph{LLC}), 13(3), 111--117. \url{https://doi.org/10.1093/llc/13.3.111}

Harrison, P. (2002). Another Addition to the \emph{An Shigao} Corpus? Preliminary Notes on an Early Chinese \emph{Saṃyukta-āgama} Translation. In \emph{Early Buddhism and Abhidharma Thought: In Honour of Dr. Hajime Sakurabe on His Seventy-seventh Birthday} (pp. 1--32). Kyoto: Heirakuji Shoten.

Hayashiya, T.\zh{（林屋友次郎）}. (1941). \emph{Kyōroku kenkyū (zenpen) \zh{經録研究（前篇）}.} Tokyo: Iwanami Shoten.

Hung, J.-J., Hsieh, C.-E., \& Bingenheimer, M. (2017). Stylometric Analysis of Chinese Buddhist Texts---Do Different Chinese Translations of the \emph{Gaṇḍavyūha} Reflect Stylistic Features that are Typical for Their Age? \emph{Journal of the Japanese Association for Digital Humanities}, 2(1), 1--30. \url{https://doi.org/10.17928/jjadh.2.1_1}

Jiayan Toolkit. \emph{Jiayan: NLP toolkit for Classical Chinese} {[}Software{]}. GitHub. \href{https://github.com/jiaeyan/Jiayan}{https://github.com/jiaeyan/Jiayan}

Lin, Y.-M. (2010). \emph{A Study on the Anthology Za Ahan Jing (T101): Centered on its Linguistic Features, Translation Style, Authorship and School Affiliation.} Saarbrücken: LAP LAMBERT Academic Publishing. ISBN 978-3843379861.

Lu, L., Zheng, Y., \& Fang, Y. (2025). Utilizing Language Models for the Attribution of Chinese Buddhist Translations: A Case Study of An Shigao's Works. \emph{Journal of Zhejiang University (Humanities and Social Sciences)}, 55(2). \url{https://doi.org/10.3785/j.issn.1008-942X.CN33-6000/C.2024.03.061}

Nattier, J. (2008). \emph{A Guide to the Earliest Chinese Buddhist Translations: Texts from the Eastern Han and Three Kingdoms Periods.} Tokyo: IRIAB, Soka University. ISBN 978-4904234006.

Zacchetti, S. (2010). Defining \emph{An Shigao}'s Translation Corpus: The State of the Art in Relevant Research. In Shen, W.-R. (Ed.), \emph{Historical and Philological Studies of China's Western Regions} (No. 3), 249--270. Beijing: Kexue Chubanshe.

Zürcher, E. (1991). A New Look at the Earliest Chinese Buddhist Texts. In K. Shinohara \& G. Schopen (Eds.), \emph{From Benares to Beijing: Essays on Buddhism and Chinese Religion in Honour of Prof. Jan Yün-hua} (pp. 277--304). Oakville, Ontario: Mosaic Press. ISBN 978-0889624443.

Bingenheimer, M. (2020). A Study and Translation of the \emph{Yakṣa-saṃyukta} in the Shorter Chinese \emph{Saṃyukta-āgama}. In Dhammadinnā (Ed.), \emph{Research on the Saṃyukta-āgama} (pp. 763--841). Taipei: Dharma Drum Publishing Corporation. ISBN 978-957-598-859-3.

\clearpage
\section{Appendix A. Dataset Composition}

\textbf{Table A1. Authenticated translations and their metadata (CBETA)}
\par\nopagebreak
\begingroup
\fontsize{9.4}{11}\selectfont\setlength{\tabcolsep}{5pt}\renewcommand{\arraystretch}{.94}
\begin{longtable}[]{@{}
  >{\raggedright\arraybackslash}p{(\columnwidth - 10\tabcolsep) * \real{0.1481}}
  >{\raggedright\arraybackslash}p{(\columnwidth - 10\tabcolsep) * \real{0.4019}}
  >{\raggedright\arraybackslash}p{(\columnwidth - 10\tabcolsep) * \real{0.0969}}
  >{\raggedright\arraybackslash}p{(\columnwidth - 10\tabcolsep) * \real{0.1071}}
  >{\raggedright\arraybackslash}p{(\columnwidth - 10\tabcolsep) * \real{0.1207}}
  >{\raggedright\arraybackslash}p{(\columnwidth - 10\tabcolsep) * \real{0.1254}}@{}}
\toprule\noalign{}
\begin{minipage}[b]{\linewidth}\raggedright
Translator
\end{minipage} & \begin{minipage}[b]{\linewidth}\raggedright
Title (English)
\end{minipage} & \begin{minipage}[b]{\linewidth}\raggedright
ID
\end{minipage} & \begin{minipage}[b]{\linewidth}\raggedright
Volumes
\end{minipage} & \begin{minipage}[b]{\linewidth}\raggedright
Sentences
\end{minipage} & \begin{minipage}[b]{\linewidth}\raggedright
Characters
\end{minipage} \\
\midrule\noalign{}
\endhead
\bottomrule\noalign{}
\endlastfoot
Yan Fotiao & Six Pāramitās Sutra & T0778 & 1 & 52 & 1069 \\
An Shigao & Long Āgama on Ten Dharmas & T0013 & 2 & 574 & 10090 \\
& Sūtra on the Desire for Human Life & T0014 & 1 & 529 & 6255 \\
& Sūtra on the Four Noble Truths & T0031 & 1 & 101 & 1858 \\
& Sūtra on the Roots of Suffering & T0032 & 1 & 239 & 3655 \\
& Sūtra on the Four Truths of Suffering & T0036 & 1 & 32 & 729 \\
& Sūtra on the True Dharma & T0048 & 1 & 44 & 1330 \\
& Discourse on Giving & T0057 & 1 & 126 & 2760 \\
& Prajñā Sūtra & T0098 & 1 & 125 & 3576 \\
& Eightfold Path Sūtra & T0112 & 1 & 44 & 626 \\
& Seven Contemplations Sūtra & T0150A & 1 & 602 & 10472 \\
& Nine Reflections Sūtra & T0150B & 1 & 17 & 489 \\
& Mindfulness Sūtra & T0603 & 2 & 309 & 5772 \\
& Earth Sūtra & T0607 & 1 & 198 & 6529 \\
Kang Senghui & Sūtra on the Six Pāramitās & T0152 & 8 & 3308 & 56217 \\
Kang Mengxiang & Sūtra on the Four Tours of the Buddha & T0137 & 1 & 58 & 1166 \\
& Sūtra on the Rise of the Buddha & T0197 & 2 & 661 & 13537 \\
Lokakṣema (Zhi Loujiachan) & Sūtra of the Bodhisattva Path & T0224 & 10 & 2773 & 58944 \\
& Sūtra on Ghosts and Spirits & T0280 & 1 & 68 & 2145 \\
& Sūtra on the Samādhi of the Vessel & T0417 & 1 & 317 & 6415 \\
Zhi Qian & Sūtra of Four Sons & T0054 & 1 & 75 & 1651 \\
& Sūtra of the Brahmin and the Demons & T0068 & 1 & 216 & 4617 \\
& Sūtra on the Transformation of Merit & T0076 & 1 & 141 & 3965 \\
& Sūtra of Prince Sudhana's Awakening & T0185 & 2 & 476 & 8370 \\
& Sūtra of the Bodhisattva's Karma & T0281 & 1 & 194 & 5285 \\
& Sūtra of Vaiśravaṇa & T0474 & 2 & 1137 & 22930 \\
Dharmarakṣa (Zhu Fahu) & Sūtra of Life & T0154 & 5 & 2398 & 48319 \\
& Sūtra on Universal Light & T0186 & 8 & 2136 & 58907 \\
& Sūtra on Radiant Light & T0222 & 10 & 3872 & 82324 \\
& Sūtra on the True Dharma & T0263 & 10 & 2833 & 76382 \\
& Mahāparinirvāṇa Sūtra & T0398 & 8 & 1379 & 38580 \\
\end{longtable}
\endgroup

\textbf{Table A2. Eight disputed translations of An Shigao and their metadata (CBETA)}
\par\nopagebreak
\begingroup
\fontsize{9.4}{11}\selectfont\setlength{\tabcolsep}{5pt}\renewcommand{\arraystretch}{.94}
\begin{longtable}[]{@{}
  >{\raggedright\arraybackslash}p{(\columnwidth - 8\tabcolsep) * \real{0.4965}}
  >{\raggedright\arraybackslash}p{(\columnwidth - 8\tabcolsep) * \real{0.1334}}
  >{\raggedright\arraybackslash}p{(\columnwidth - 8\tabcolsep) * \real{0.1120}}
  >{\raggedright\arraybackslash}p{(\columnwidth - 8\tabcolsep) * \real{0.1266}}
  >{\raggedright\arraybackslash}p{(\columnwidth - 8\tabcolsep) * \real{0.1315}}@{}}
\toprule\noalign{}
\begin{minipage}[b]{\linewidth}\raggedright
Title (English)
\end{minipage} & \begin{minipage}[b]{\linewidth}\raggedright
CBETA ID
\end{minipage} & \begin{minipage}[b]{\linewidth}\raggedright
Volumes
\end{minipage} & \begin{minipage}[b]{\linewidth}\raggedright
Sentences
\end{minipage} & \begin{minipage}[b]{\linewidth}\raggedright
Characters
\end{minipage} \\
\midrule\noalign{}
\endhead
\bottomrule\noalign{}
\endlastfoot
Saṃyuktāgama Sutra & T0101 & 1 & 428 & 9247 \\
Sūtra on the Five Aggregates & T0105 & 1 & 51 & 787 \\
Dhammacakkappavattana Sutta & T0109 & 1 & 32 & 770 \\
Thirty-Seven Factors of Enlightenment & T0604 & 1 & 16 & 931 \\
Sūtra on Meditation and Contemplation & T0605 & 1 & 10 & 281 \\
Sūtra on Receiving the Dharma & T0792 & 1 & 17 & 304 \\
Abhidharma Prakaraṇa Sūtra & T1557 & 1 & 385 & 4368 \\
Mahāsaṃnipāta Sūtra, Chapter on Ten-Direction Bodhisattvas & T0397\_059 & 1 & 316 & 9809 \\
\end{longtable}
\endgroup

\clearpage
\section{Appendix B. Feature System for Stylometric Analysis}

\textbf{Table B1. Sentence-final Character Features}
\par\nopagebreak
\begingroup
\small\setlength{\tabcolsep}{5pt}
\begin{longtable}[]{@{}
  >{\raggedright\arraybackslash}p{(\columnwidth - 4\tabcolsep) * \real{0.2265}}
  >{\raggedright\arraybackslash}p{(\columnwidth - 4\tabcolsep) * \real{0.3681}}
  >{\raggedright\arraybackslash}p{(\columnwidth - 4\tabcolsep) * \real{0.4054}}@{}}
\toprule\noalign{}
\begin{minipage}[b]{\linewidth}\raggedright
Feature Type
\end{minipage} & \begin{minipage}[b]{\linewidth}\raggedright
Characters (Chinese)
\end{minipage} & \begin{minipage}[b]{\linewidth}\raggedright
Transliteration
\end{minipage} \\
\midrule\noalign{}
\endhead
\bottomrule\noalign{}
\endlastfoot
Sentence-final Char & \zh{也}, \zh{是}, \zh{之}, \zh{乎}, \zh{者}, \zh{故}, \zh{何}, \zh{行}, \zh{提}, \zh{法} & \emph{ye, shi, zhi, hu, zhe, gu, he, xing, ti, fa} \\
\end{longtable}
\endgroup

\textbf{Table B2. Sentence-formula Features}
\par\nopagebreak
\begingroup
\small\setlength{\tabcolsep}{5pt}
\begin{longtable}[]{@{}
  >{\raggedright\arraybackslash}p{(\columnwidth - 4\tabcolsep) * \real{0.2537}}
  >{\raggedright\arraybackslash}p{(\columnwidth - 4\tabcolsep) * \real{0.3421}}
  >{\raggedright\arraybackslash}p{(\columnwidth - 4\tabcolsep) * \real{0.4042}}@{}}
\toprule\noalign{}
\begin{minipage}[b]{\linewidth}\raggedright
Template
\end{minipage} & \begin{minipage}[b]{\linewidth}\raggedright
Example (Chinese)
\end{minipage} & \begin{minipage}[b]{\linewidth}\raggedright
Function (English)
\end{minipage} \\
\midrule\noalign{}
\endhead
\bottomrule\noalign{}
\endlastfoot
\zh{第} X & \zh{第一、第二}\ldots{} & Sequential enumeration \\
X \zh{者} & \zh{色者、心者}\ldots{} & Definition or exemplification \\
\zh{以} X \zh{为} Y & \zh{以心为界、以苦为道} & Categorical relation \\
\zh{何等故举} & \zh{何等、何者、云何} & Interrogative / doctrinal question \\
\zh{佛言} & \zh{佛言：}\ldots{} & Quotation formula \\
\zh{已} X \zh{便} & \zh{已灭便得道} & Temporal / conditional sequence \\
\end{longtable}
\endgroup

\textbf{Table B3. Function-characters Features}
\par\nopagebreak
\begingroup
\small\setlength{\tabcolsep}{5pt}
\begin{longtable}[]{@{}
  >{\raggedright\arraybackslash}p{(\columnwidth - 4\tabcolsep) * \real{0.1916}}
  >{\raggedright\arraybackslash}p{(\columnwidth - 4\tabcolsep) * \real{0.2835}}
  >{\raggedright\arraybackslash}p{(\columnwidth - 4\tabcolsep) * \real{0.5250}}@{}}
\toprule\noalign{}
\begin{minipage}[b]{\linewidth}\raggedright
Category
\end{minipage} & \begin{minipage}[b]{\linewidth}\raggedright
Function Words (Chinese)
\end{minipage} & \begin{minipage}[b]{\linewidth}\raggedright
English Gloss
\end{minipage} \\
\midrule\noalign{}
\endhead
\bottomrule\noalign{}
\endlastfoot
Pronouns & \zh{之}, \zh{其}, \zh{自}, \zh{皆}, \zh{是} & pronouns (of, its, self, all, to be) \\
Prepositions & \zh{以}, \zh{於}, \zh{为}, \zh{与}, \zh{及} & prepositions (by, in, for, with, and) \\
Particles & \zh{所}, \zh{者}, \zh{故} & particles (that which, one who, therefore) \\
Conjunctions / Adverbs & \zh{而}, \zh{有}, \zh{则}, \zh{无}, \zh{不}, \zh{如}, \zh{能} & conjunctions/adverbs (and, have, then, none, not, like, can) \\
\end{longtable}
\endgroup

\textbf{Table B4. Top-10 most discriminative stylometric features across translators (Kruskal--Wallis \emph{p} \textless{} 0.01)}
\par\nopagebreak
\begingroup
\small\setlength{\tabcolsep}{5pt}
\begin{longtable}[]{@{}
  >{\raggedright\arraybackslash}p{(\columnwidth - 8\tabcolsep) * \real{0.2165}}
  >{\raggedright\arraybackslash}p{(\columnwidth - 8\tabcolsep) * \real{0.5721}}
  >{\raggedright\arraybackslash}p{(\columnwidth - 8\tabcolsep) * \real{0.1228}}
  >{\raggedright\arraybackslash}p{(\columnwidth - 8\tabcolsep) * \real{0.0148}}
  >{\raggedright\arraybackslash}p{(\columnwidth - 8\tabcolsep) * \real{0.0737}}@{}}
\toprule\noalign{}
\begin{minipage}[b]{\linewidth}\raggedright
Feature
\end{minipage} & \begin{minipage}[b]{\linewidth}\raggedright
Description
\end{minipage} & \multicolumn{2}{>{\raggedright\arraybackslash}p{(\columnwidth - 8\tabcolsep) * \real{0.1376} + 2\tabcolsep}}{%
\begin{minipage}[b]{\linewidth}\raggedright
p\_holm
\end{minipage}} & \begin{minipage}[b]{\linewidth}\raggedright
\ensuremath{\varepsilon^2}
\end{minipage} \\
\midrule\noalign{}
\endhead
\bottomrule\noalign{}
\endlastfoot
func\_\zh{之}\_per100 & Frequency of particle zhi (\zh{之},nominalizer) per 100 chars & \(4.80\times10^{-2}\) & \multicolumn{2}{>{\raggedright\arraybackslash}p{(\columnwidth - 8\tabcolsep) * \real{0.0885} + 2\tabcolsep}@{}}{%
0.838} \\
func\_\zh{其}\_per100 & Frequency of pronoun qi (\zh{其},``his/her/its'') per 100 chars & \(7.22\times10^{-2}\) & \multicolumn{2}{>{\raggedright\arraybackslash}p{(\columnwidth - 8\tabcolsep) * \real{0.0885} + 2\tabcolsep}@{}}{%
0.798} \\
pos3\_v\_u\_n\_prop & POS trigram Verb--Auxiliary--Noun proportion & \(1.19\times10^{-1}\) & \multicolumn{2}{>{\raggedright\arraybackslash}p{(\columnwidth - 8\tabcolsep) * \real{0.0885} + 2\tabcolsep}@{}}{%
0.749} \\
func\_\zh{皆}\_per100 & Frequency of adverb jie (\zh{皆}, ``all'') per 100 chars & \(1.33\times10^{-1}\) & \multicolumn{2}{>{\raggedright\arraybackslash}p{(\columnwidth - 8\tabcolsep) * \real{0.0885} + 2\tabcolsep}@{}}{%
0.738} \\
func\_\zh{及}\_per100 & Frequency of conjunction ji (\zh{及}, ``and '') per 100 chars & \(1.36\times10^{-1}\) & \multicolumn{2}{>{\raggedright\arraybackslash}p{(\columnwidth - 8\tabcolsep) * \real{0.0885} + 2\tabcolsep}@{}}{%
0.735} \\
pos3\_d\_v\_d\_prop & POS trigram Adverb--Verb--Adverb proportion & \(1.52\times10^{-1}\) & \multicolumn{2}{>{\raggedright\arraybackslash}p{(\columnwidth - 8\tabcolsep) * \real{0.0885} + 2\tabcolsep}@{}}{%
0.724} \\
pos3\_n\_u\_v\_prop & POS trigram Noun--Auxiliary--Verb proportion & \(1.66\times10^{-1}\) & \multicolumn{2}{>{\raggedright\arraybackslash}p{(\columnwidth - 8\tabcolsep) * \real{0.0885} + 2\tabcolsep}@{}}{%
0.715} \\
pos4\_n\_v\_r\_n\_prop & POS 4-gram Noun--Verb--Particle--Noun proportion & \(1.88\times10^{-1}\) & \multicolumn{2}{>{\raggedright\arraybackslash}p{(\columnwidth - 8\tabcolsep) * \real{0.0885} + 2\tabcolsep}@{}}{%
0.702} \\
pos2\_v\_d\_prop & POS bigram Verb--Adverb proportion & \(1.88\times10^{-1}\) & \multicolumn{2}{>{\raggedright\arraybackslash}p{(\columnwidth - 8\tabcolsep) * \real{0.0885} + 2\tabcolsep}@{}}{%
0.702} \\
pos2\_n\_r\_prop & POS bigram Noun--Particle proportion & \(2.02\times10^{-1}\) & \multicolumn{2}{>{\raggedright\arraybackslash}p{(\columnwidth - 8\tabcolsep) * \real{0.0885} + 2\tabcolsep}@{}}{%
0.694} \\
\end{longtable}
\endgroup

\emph{Note:} \(p_{holm}\)= Holm--Bonferroni corrected significance level;\(\varepsilon^{2}\) = effect size representing the proportion of variance in feature ranks explained by translator differences.

\clearpage
\section{Appendix C. Results Details}

\textbf{Table C1. Stylometric scores distribution across each scripture translation}
\par\nopagebreak
\begingroup
\fontsize{8.2}{10}\selectfont\setlength{\tabcolsep}{2.4pt}\renewcommand{\arraystretch}{.88}
\begin{longtable}{@{}p{69pt}l*{10}{r}@{}}
\toprule
Author & Scripture & \shortstack{func\_\zh{之}\\\_per100} & \shortstack{func\_\zh{其}\\\_per100} & \shortstack{func\_\zh{皆}\\\_per100} & \shortstack{func\_\zh{及}\\\_per100} & \shortstack{pos2\_\\v\_d\\\_prop} & \shortstack{pos2\_\\n\_r\\\_prop} & \shortstack{pos3\_\\v\_u\_n\\\_prop} & \shortstack{pos3\_\\d\_v\_d\\\_prop} & \shortstack{pos3\_\\n\_u\_v\\\_prop} & \shortstack{pos4\_\\n\_v\_r\_n\\\_prop} \\
\midrule
\endfirsthead
\toprule
Author & Scripture & \shortstack{func\_\zh{之}\\\_per100} & \shortstack{func\_\zh{其}\\\_per100} & \shortstack{func\_\zh{皆}\\\_per100} & \shortstack{func\_\zh{及}\\\_per100} & \shortstack{pos2\_\\v\_d\\\_prop} & \shortstack{pos2\_\\n\_r\\\_prop} & \shortstack{pos3\_\\v\_u\_n\\\_prop} & \shortstack{pos3\_\\d\_v\_d\\\_prop} & \shortstack{pos3\_\\n\_u\_v\\\_prop} & \shortstack{pos4\_\\n\_v\_r\_n\\\_prop} \\
\midrule
\endhead
\bottomrule
\endfoot
Yan Fotiao & T0778 & 41.10\% & 28.47\% & 46.27\% & 29.34\% & 37.14\% & 31.10\% & 50.81\% & 39.18\% & 24.77\% & 23.18\% \\
An Shigao & T0013 & 29.07\% & 28.47\% & 31.49\% & 31.06\% & 71.76\% & 28.34\% & 30.59\% & 74.50\% & 23.47\% & 27.04\% \\
& T0014 & 29.07\% & 28.47\% & 31.54\% & 29.34\% & 56.83\% & 37.28\% & 27.95\% & 63.44\% & 30.22\% & 34.96\% \\
& T0031 & 29.07\% & 28.47\% & 30.85\% & 29.34\% & 89.42\% & 47.01\% & 22.98\% & 94.63\% & 29.36\% & 56.40\% \\
& T0032 & 29.07\% & 29.46\% & 32.40\% & 29.34\% & 82.59\% & 23.21\% & 26.95\% & 72.50\% & 27.32\% & 26.67\% \\
& T0036 & 29.07\% & 28.47\% & 30.85\% & 29.34\% & 66.03\% & 20.82\% & 22.98\% & 60.89\% & 17.11\% & 23.18\% \\
& T0048 & 29.07\% & 28.47\% & 30.85\% & 29.34\% & 56.20\% & 32.37\% & 22.98\% & 75.20\% & 17.11\% & 23.18\% \\
& T0057 & 29.07\% & 28.47\% & 30.85\% & 29.34\% & 92.12\% & 20.82\% & 28.62\% & 78.34\% & 19.60\% & 23.18\% \\
& T0098 & 29.07\% & 28.47\% & 33.84\% & 29.34\% & 77.20\% & 22.05\% & 36.72\% & 74.74\% & 27.83\% & 26.83\% \\
& T0112 & 29.07\% & 28.47\% & 39.69\% & 29.34\% & 62.24\% & 60.91\% & 37.12\% & 77.30\% & 49.02\% & 50.09\% \\
& T0150A & 30.03\% & 29.64\% & 33.33\% & 30.98\% & 57.49\% & 35.65\% & 32.35\% & 59.69\% & 44.95\% & 32.87\% \\
& T0150B & 29.07\% & 28.47\% & 30.85\% & 29.34\% & 90.58\% & 20.82\% & 22.98\% & 79.70\% & 34.60\% & 23.18\% \\
& T0603 & 29.69\% & 28.47\% & 33.26\% & 29.34\% & 75.66\% & 28.88\% & 30.77\% & 76.69\% & 25.87\% & 35.11\% \\
& T0607 & 29.07\% & 28.47\% & 36.63\% & 40.01\% & 40.94\% & 34.02\% & 26.57\% & 46.82\% & 38.03\% & 29.76\% \\
Lokakṣema & T0224 & 46.09\% & 57.73\% & 51.25\% & 59.22\% & 40.15\% & 64.21\% & 93.69\% & 43.25\% & 56.12\% & 42.04\% \\
& T0280 & 31.94\% & 94.08\% & 98.56\% & 45.29\% & 27.38\% & 97.86\% & 39.66\% & 32.28\% & 90.10\% & 54.77\% \\
& T0417 & 56.05\% & 73.50\% & 51.40\% & 76.73\% & 44.03\% & 65.97\% & 56.31\% & 53.19\% & 66.19\% & 73.55\% \\
Zhi Qian & T0054 & 61.33\% & 57.31\% & 87.32\% & 29.34\% & 31.44\% & 49.85\% & 51.59\% & 54.16\% & 60.63\% & 81.60\% \\
& T0068 & 61.66\% & 52.03\% & 68.49\% & 55.16\% & 35.25\% & 54.11\% & 63.11\% & 38.49\% & 57.14\% & 52.36\% \\
& T0076 & 92.93\% & 70.20\% & 57.07\% & 29.34\% & 30.63\% & 36.31\% & 78.03\% & 29.91\% & 84.56\% & 57.84\% \\
& T0185 & 63.69\% & 65.90\% & 60.22\% & 64.56\% & 36.08\% & 56.95\% & 67.89\% & 38.31\% & 62.58\% & 60.64\% \\
& T0281 & 56.21\% & 38.91\% & 62.61\% & 58.18\% & 24.32\% & 51.56\% & 58.79\% & 16.64\% & 63.59\% & 41.83\% \\
& T0474 & 62.34\% & 69.91\% & 59.78\% & 60.86\% & 39.22\% & 69.31\% & 64.37\% & 41.67\% & 58.66\% & 68.66\% \\
Kang Senghui & T0152 & 87.55\% & 68.88\% & 42.92\% & 41.92\% & 31.83\% & 53.82\% & 86.78\% & 28.18\% & 58.39\% & 73.03\% \\
Kang Mengxiang & T0137 & 36.04\% & 56.41\% & 37.47\% & 82.46\% & 25.33\% & 65.30\% & 48.04\% & 15.28\% & 90.45\% & 94.17\% \\
& T0197 & 47.94\% & 45.09\% & 65.52\% & 87.66\% & 29.00\% & 57.32\% & 41.18\% & 30.61\% & 67.66\% & 74.63\% \\
Dharmarakṣa & T0154 & 77.08\% & 73.08\% & 53.53\% & 80.07\% & 35.28\% & 66.23\% & 60.52\% & 33.17\% & 54.90\% & 80.14\% \\
& T0186 & 70.49\% & 77.27\% & 73.62\% & 86.61\% & 29.95\% & 67.92\% & 51.31\% & 28.54\% & 59.57\% & 74.02\% \\
& T0222 & 56.91\% & 57.02\% & 43.77\% & 75.18\% & 45.74\% & 72.51\% & 90.70\% & 41.77\% & 55.97\% & 62.38\% \\
& T0263 & 80.18\% & 78.50\% & 71.62\% & 89.11\% & 30.68\% & 69.20\% & 58.98\% & 28.58\% & 72.41\% & 74.82\% \\
& T0398 & 78.41\% & 89.28\% & 53.07\% & 77.98\% & 35.01\% & 70.26\% & 57.20\% & 29.93\% & 77.29\% & 66.11\% \\
\end{longtable}
\endgroup

\textbf{Table C2. Aggregated stylometric scores distribution across each translator}
\par\nopagebreak
\begingroup
\fontsize{8.2}{10}\selectfont\setlength{\tabcolsep}{2.4pt}\renewcommand{\arraystretch}{.88}
\begin{longtable}{@{}p{85pt}*{10}{r}@{}}
\toprule
Author & \shortstack{func\_\zh{之}\\\_per100} & \shortstack{func\_\zh{其}\\\_per100} & \shortstack{func\_\zh{皆}\\\_per100} & \shortstack{func\_\zh{及}\\\_per100} & \shortstack{pos2\_\\v\_d\\\_prop} & \shortstack{pos2\_\\n\_r\\\_prop} & \shortstack{pos3\_\\v\_u\_n\\\_prop} & \shortstack{pos3\_\\d\_v\_d\\\_prop} & \shortstack{pos3\_\\n\_u\_v\\\_prop} & \shortstack{pos4\_\\n\_v\_r\_n\\\_prop} \\
\midrule
\endfirsthead
\toprule
Author & \shortstack{func\_\zh{之}\\\_per100} & \shortstack{func\_\zh{其}\\\_per100} & \shortstack{func\_\zh{皆}\\\_per100} & \shortstack{func\_\zh{及}\\\_per100} & \shortstack{pos2\_\\v\_d\\\_prop} & \shortstack{pos2\_\\n\_r\\\_prop} & \shortstack{pos3\_\\v\_u\_n\\\_prop} & \shortstack{pos3\_\\d\_v\_d\\\_prop} & \shortstack{pos3\_\\n\_u\_v\\\_prop} & \shortstack{pos4\_\\n\_v\_r\_n\\\_prop} \\
\midrule
\endhead
\bottomrule
\endfoot
Yan Fotiao & 41.10\% & 28.47\% & 46.27\% & 29.34\% & 37.14\% & 31.10\% & 50.81\% & 39.18\% & 24.77\% & 23.18\% \\
An Shigao & 29.19\% & 28.64\% & 32.80\% & 30.42\% & 70.70\% & 31.71\% & 28.43\% & 71.88\% & 29.58\% & 31.73\% \\
Kang Senghui & 87.55\% & 68.88\% & 42.92\% & 41.92\% & 31.83\% & 53.82\% & 86.78\% & 28.18\% & 58.39\% & 73.03\% \\
Kang Mengxiang & 41.99\% & 50.75\% & 51.49\% & 85.06\% & 27.17\% & 61.31\% & 44.61\% & 22.94\% & 79.05\% & 84.40\% \\
Lokakṣema & 44.70\% & 75.10\% & 67.07\% & 60.41\% & 37.19\% & 76.01\% & 63.22\% & 42.91\% & 70.80\% & 56.79\% \\
Zhi Qian & 66.36\% & 59.04\% & 65.91\% & 49.57\% & 32.82\% & 53.02\% & 63.96\% & 36.53\% & 64.53\% & 60.49\% \\
Dharmarakṣa & 72.61\% & 75.03\% & 59.12\% & 81.79\% & 35.33\% & 69.22\% & 63.74\% & 32.40\% & 64.03\% & 71.49\% \\
\end{longtable}
\endgroup

\textbf{Table C3. Classification of Disputed An Shigao Translations Results}
\par\nopagebreak
\begingroup
\fontsize{9.4}{11}\selectfont\setlength{\tabcolsep}{5pt}\renewcommand{\arraystretch}{.94}
\begin{longtable}[]{@{}
  >{\raggedright\arraybackslash}p{(\columnwidth - 4\tabcolsep) * \real{0.1352}}
  >{\raggedright\arraybackslash}p{(\columnwidth - 4\tabcolsep) * \real{0.3451}}
  >{\raggedright\arraybackslash}p{(\columnwidth - 4\tabcolsep) * \real{0.5197}}@{}}
\toprule\noalign{}
\begin{minipage}[b]{\linewidth}\raggedright
CBETA ID
\end{minipage} & \begin{minipage}[b]{\linewidth}\raggedright
Our Results
\end{minipage} & \begin{minipage}[b]{\linewidth}\raggedright
BERT model Results
\end{minipage} \\
\midrule\noalign{}
\endhead
\bottomrule\noalign{}
\endlastfoot
T0101 & Yes & Yes \\
T0105 & No & No \\
T0109 & No & No \\
T0604 & No & No \\
T0605 & No & No \\
T0792 & No & No \\
T1557 & An Shigao & An Shigao \\
T0397\_059 & No & No \\
\end{longtable}
\endgroup

\end{document}